\pdfoutput=1

\documentclass[11pt]{article}

\usepackage[final]{acl}
\usepackage{times}
\usepackage{latexsym}

\usepackage[T1]{fontenc}

\usepackage[utf8]{inputenc}

\usepackage{microtype}

\usepackage{inconsolata}

\usepackage{graphicx}
\usepackage{amsmath}
\usepackage{subcaption}
\usepackage{geometry}
\usepackage{tabularx}
\usepackage{array}
\usepackage{caption}
\usepackage{times}
\usepackage{latexsym}
\usepackage{lipsum}
\usepackage{tcolorbox}
\usepackage{booktabs}
\usepackage{caption}

\title{{\em Pron vs Prompt}: Can Large Language Models already Challenge a World-Class Fiction Author at Creative Text Writing?}

\author{
 \textbf{Guillermo Marco\textsuperscript{\dag}},
 \textbf{Julio Gonzalo\textsuperscript{\dag}},
 \textbf{Teresa Mateo\textsuperscript{\ddag}}
 \textbf{Ramón del Castillo\textsuperscript{\S}}
\\
\\
 \textsuperscript{\dag} School of Computer Science, UNED, Madrid, Spain
 \\
 \textsuperscript{\ddag}Faculty of Education, UCM, Madrid, Spain
\\
 \textsuperscript{\S}Faculty of Philosophy, UNED, Madrid, Spain
\\
 \small{
   \textbf{Correspondence:} \href{gmarco@lsi.uned.es}{gmarco@lsi.uned.es}
 }
}

\begin{document}
\maketitle
\begin{abstract}
Are LLMs ready to compete in creative writing skills with a top (rather than average) novelist? To provide an initial answer for this question, we have carried out a contest between Patricio Pron (an awarded novelist, considered one of the best of his generation) and GPT-4 (one of the top performing LLMs), in the spirit of AI-human duels such as DeepBlue vs Kasparov and AlphaGo vs Lee Sidol. We asked Pron and GPT-4 to provide thirty titles each, and then to write short stories for both their titles and their opponent's. Then, we prepared an evaluation rubric inspired by Boden's definition of creativity, and we collected several detailed expert assessments of the texts, provided by literature critics and scholars. The results of our experimentation indicate that LLMs are still far from challenging a top human creative writer. We also observed that GPT-4 writes more creatively using Pron's titles than its own titles (which is an indication of the potential for human-machine co-creation). Additionally, we found that GPT-4 has a more creative writing style in English than in Spanish. 


\end{abstract}

\section{Introduction}

Large Language Models (LLMs) have recently showed strong competences generating human-like text, and in particular in creative writing tasks \cite{achiam2023gpt}, which is the focus of this paper. LLMs are increasingly influencing creative industries, impacting both the economy and the labor market, as highlighted by significant events such as the Hollywood screenwriters' strike \cite{lee_industries_2022, eloundou2023gpts, koblin_whats_2023}. Experimentation shows that, under different settings, LLMs can perform better than average humans at short creative writing tasks \cite{marco_transformers_2023,gomez-rodriguez-williams-2023-confederacy}. 

LLMs seem to be ready, then, for the next level of experimental inquiry: {\em Can they already compete with the best human creative writers?} Note that, in the history of Artificial Intelligence (AI), symbolic landmarks involve competition between the best AI systems with the best humans at the task, as in DeepBlue vs Kasparov \cite{campbell2002deep} and AlphaGo vs Lee Sidol \cite{silver2016mastering}. However, despite extensive research into human-machine collaboration \cite{fang2023systematic,li_value_2024}, there is still little experimentation on how the best LLMs compare with the best fiction writers in autonomous creative text writing.  

In this work, we make the first attempt (known to us) to conduct a formal contest of autonomous creative writing between two top writers: GPT-4 Turbo\footnote{GPT-4 for short in the remainder of the paper} (\texttt{gpt-4-0125-preview}, the best LLM at the time of conducting this research, together with Claude-3 Opus \cite{claude_2024} and Gemini Ultra \cite{team2023gemini}) and Patricio Pron\footnote{\url{https://en.wikipedia.org/wiki/Patricio_Pron}}, a distinguished Spanish-speaking writer recognized among the 22 best writers of his generation by Granta Magazine\footnote{\url{https://www.nytimes.com/2010/10/02/books/02granta.html}}.

Our experimentation, and this paper, are structured along the following research questions: 

\textbf{RQ1: Can the current state of generative AI match the skills of the best human authors in creative writing tasks?} 

As in previous AI duels, we do not try to compare a top AI machine with average humans: we focus on a one-on-one comparison between two (AI and human) top performers. Comparing with one top writer certainly limits the scope of our results, but also lets us put all experimental efforts in providing a comprehensive side by side evaluation, which involves designing 60 text writing assignments, under the same conditions for both contenders, and collecting the manual evaluation of literary critics on those 180 pieces of text (60 by Pron, 60 by GPT-4 in English and 60 by GPT-4 in Spanish), with a carefully crafted rubric composed of 10 questions each (see Section \ref{sec:exp-design} on methodology).

\textbf{RQ2: What is the role of the prompt in the creativity of the generated text?} 

Different studies indicate that through prompting the model can be guided to write more creative texts \cite{bellemare-pepin_divergent_2024}. In our experiment, each text assignment is just a proposed title, and titles are provided by GPT-4 and Patricio Pron (30 titles each). In a second stage, they have to write 60 synopsis of imaginary movies with the proposed title, both for their own titles and their contender's. This lets us explore how the source of the title influences the quality dimensions of the texts produced, for each of the authors. By "prompt" we do not mean how things are asked to the LLM; in our case the prompt variations consist in what is asked ---the requested topic of the text as conveyed by the title. We believe that, in a creative writing setting, the request is part of the creative process, and we want to model how it influences the result. 

\textbf{RQ3: Are LLMs less skilled for creative writing in languages other than English?}

The largest LLMs are often trained on unbalanced training data across languages where English is predominant. For example, Llama-3 \cite{meta_llama3_2024} is only trained on 5\% multilingual data. In our experimentation, we compare the performance of GPT-4 in English with its performance in Spanish, the third most spoken language in the world, to verify if there is a performance gap even with respect to other resource-rich languages. 

\textbf{RQ4: Does GPT-4 have a recognizable style for a literature expert when generating creative text without constraints?}

 In our experimentation, we do not constraint GPT-4 writing style via prompting, and we request assessors to (blindly) identify if each text has been written by a machine or a human. We then explore not only if the assessors are able to recognize machine-produce text, but whether this recognition improves along the evaluation process. 

\textbf{RQ5: Can we effectively measure creativity using Boden's framework in the context of AI-generated texts?}

Margaret Boden's definition of creativity \cite{boden2004,boden2010creativity} requires novelty, surprise, and value in creative outputs. These minimal requisites are ubiquitous in most definitions of creativity, and we used them as a guide to prepare an evaluation rubric for our expert assessors. The rubric provides a framework for an objective analysis of Boden's dimensions, where we can measure if originality (novelty and surprise in the context of literary writing) and attractiveness (the value of the literary text) do correlate with creativity assessments.  

The main contributions of our work are: 

1. We conduct the first comprehensive symmetrical empirical study that compares a state of the art LLM (GPT-4) with an award-winning novelist, quoted as one of the best of its generation. With this comparison we approach the question of whether LLMs can already be better than any human at creative writing, rather than better than average humans. Our methodology includes a carefully crafted rubric to evaluate creative writing texts according to Boden's dimensions of creativity. 

2. Our results indicate that, when judged by expert critics and scholars, GPT-4 creative writing skills are not ready to compete with top human writers. The expert assessments collected strongly prefer Patricio Pron to GPT-4 in all quality dimensions considered in the study.

3. We also provide quantitative evidence that (i) prompting GPT-4 with titles provided by the novelist improves its writing; (ii) GPT-4 creative writing skills degrade in Spanish (with respect to English); and (iii) when freed from stylistic constraints, GPT-4 writing style seems to be recognizable, specially after some exposure to its writing.

\section{Related Work}

Since the rise of LLM technology, creative text writing has gained renewed interest within the NLP research community. \citet{franceschelli_creativity_2024} survey machine learning and creativity, discussing computational creativity theories, generative techniques, and evaluation methods. Evaluating creativity remains challenging \cite{hamalainen_human_2021,chakrabarty_creative_2023}, but progress is being made; for an extensive explanation of the challenges of evaluating computational creativity see  \citet{lamb_evaluating_2018}.

Regarding machine-assisted human writing, \citet{swanson_story_2021} introduced \textit{Story Centaur}, a tool for creative writers to prototype few-shot learning models. And \citet{chakrabarty_help_2022} presented CoPoet, a system for poetry writing that enhances user-generated content. In both cases, evaluators prefer texts generated in co-authorship with IA systems. However, \citet{kreminski_unmet_2022} highlighted limitations in current LLM tools, such as issues with narrative consistency and plot development.

Our focus is rather on autonomous LLMs creative writing. \citet{gunser_pure_2022} examined the stylistic quality of AI-generated texts, finding them generally rated lower than human-written texts despite being indistinguishable. \citet{marco_transformers_2023} found that a fine-tuned BART model outperformed average human writers in a creative writing task, obtaining higher scores in grammaticality, coherence and attractiveness, and almost matching their creativity. Unlike our study, they used casual readers to assess the texts, and the human texts were not produced by top writers. 


The study by \citet{gomez-rodriguez-williams-2023-confederacy} examines the capability of several large language models (LLMs) in autonomous English creative writing, focusing on a single imaginative task where models and humans compose a story about a combat between Ignatius J. Reilly and a pterodactyl. They reveal that LLMs performed well in fluency and coherence but lagged in creativity and humor. Their study's single-task focus contrasts with our broad evaluation of 60 titles. 

Lastly, \citet{chakrabarty_art_2024} proposed the Torrance Test of Creative Writing (TTCW) to evaluate AI-generated stories. Their findings reveal that while LLMs perform well in terms of fluency and structure, they lag significantly behind human writers in originality and emotional depth. A limitation of the study is that the tasks given to humans and machines are asymmetrical: human stories are selected from already published material. Then, GPT-4 summarizes the stories, and LLMs are asked to generate a full story starting from each of the summaries, which is only a part of the creative writing process. Another difference in methodology is that they adapt the TTCW test for their rubric, while we design our rubric following Boden's notion of creativity. 

Overall, our study complements previous work being the only one that simultaneously (i) uses the best possible writer and LLM for the experimentation; (ii) gives the same tasks to both contenders in equal conditions  (iii) explores 60 different writing assignments (proposed by the contenders) and collects assessments for 180 texts using a rubric that adapts Boden's notion of creativity to the task, and (iv) includes a study on the effect of the prompt and also measures the gap between texts written in English and Spanish.

\section{Experimental Design}\label{sec:exp-design}

\textbf{Contenders.} The LLM chosen for the experiment is GPT-4 Turbo (in \texttt{gpt-4-0125-preview} version), which was the strongest LLM when we initiated the experiment. After some initial experimentation with the system, we fixed temperature at 1: going beyond this value occasionally impacted on grammaticality (particularly with Spanish texts), so we chose the highest value that produced always formally correct texts. Once the experiment was initiated, other LLMs that seemed to rival the performance of GPT-4 were launched: most notably Claude 3 Opus, Gemini Ultra and Llama 3. Experimenting with these models, we did not notice any clear advantages with respect to GPT-4, so we proceeded with our initial setup. 

Finding a top novelist that would engage in this experiment was easier than we initially thought. We contacted Patricio Pron because, besides being awarded with some of the most prestigious distinctions in Spanish literature (the Alfaguara Award for Novel Writing, The Juan Rulfo narrative award, among others) and besides being translated into dozens of languages, he also has a strong curiosity towards Artificial Intelligence and autonomous machine writing.

\textbf{Task design.} In the first stage, each contender proposed 30 movie titles. In the second stage, both contenders wrote synopses (approximately 600 words) for each of the 60 titles. The prompt for GPT-4 was as follows: \textit{``We are conducting an experiment to compare your creative writing skills with those of the renowned novelist Patricio Pron. Your task is to generate synopses for imaginary movie titles. These synopses should be creative, appealing to critics and audiences, and possess inherent literary value. Here is some information about Patricio Pron: he is a celebrated writer, recognized as one of the top young writers in Spanish by Granta in 2010, and the winner of the Alfaguara Prize in 2019 for his work \textit{Mañana tendremos otros nombres}. The proposed title is: \textit{ \{title\}}. Please write a 600-word synopsis that meets these criteria.''} More information on the prompts can be found in Appendix \ref{appx:materials}.

\textbf{Languages.} Titles were originally proposed in Spanish. Then, we manually translated them into English. Pron wrote a text in Spanish for each of the titles, and GPT-4 wrote a text in Spanish and a different text in English for each of the titles, so that we can measure how its writing skills depend on the language. 

\textbf{Rubric Design.}
The rubric, designed by three experts in pedagogy, psychometrics, literature, and NLP, focuses on creativity-related dimensions, as previous work has shown that LLMs already excel at grammaticality, coherence and fluency \cite{marco_transformers_2023}. 

The point of departure is Margaret Boden's definition \cite{boden2004}: \textit{Creativity is the ability to come up with ideas that are new, surprising and valuable.}; it is a simple, operative definition compatible with most studies on the subject, both from philosophers and psychologists, with a long tradition \cite{gaut2010philosophy}. It is a conceptualization of creativity in three specific dimensions: novelty, surprise and value.

Our experts in the process of creating the rubric agreed that, in fiction writing, novelty and surprise can be conflated into one single feature, {\em originality}. They rely on Bartel's definition: a work is original if it is the first to display some unique or different attribute that is then adopted by other works \cite{lamb_evaluating_2018}. 

Value, on the other hand, is a catch-all, which involves both economic and historical dimensions of art. The approach the experts take is intrinsically product-based: they evaluate the creativity of the text in itself; regardless of historical or social considerations that would make the evaluation noisy. In the context of fiction writing, they mapped value to attractiveness: a synopsis is valuable if it engages the reader and provides a satisfying reading experience.

The experts' rubric encompasses the following quality dimensions (see Appendix \ref{apxx:rubric} for details) rated from 0 to 3:

 \textit{Attractiveness:} literary appeal of the title, the style of the text, and its content (theme/plot).  Criteria include the title's captivation, style's enjoyment, and the engagement of story and characters. 
    
    \textit{Originality:} novelty and uniqueness of the title, the text style and the text theme/plot. Criteria include the title's uniqueness, the style's distinctiveness, and the plot's innovation.
    
    \textit{Creativity:} This assessment evaluates the creativity of both the title and synopsis without distinguishing between style and theme. A unique aspect of this evaluation is the inclusion of the term "creativity" in the definition of each level. This approach aims to determine if evaluators' mental models of creativity correlate with the defined aspects of attractiveness and originality. The primary purpose of assessing creativity in this manner is to examine its correlation with attractiveness and originality. According to Boden, creativity is defined as something that is new, surprising, and valuable. Originality involves novelty and surprise regardless of the value of the text. This distinction is reflected in how experts develop the evaluation rubric, where each degree of originality is clearly defined. The creativity rubric is designed to reveal how evaluators—such as critics and literature scholars—perceive creativity. Specifically, we aim to measure its correlation with originality and attractiveness to validate Boden's definition.
    
    \textit{Anthology Potential:} Evaluates the text's fit within its genre and its potential to be included in an anthology, according to his opinion as a literary critic.

     \textit{Own voice:} evaluates if the author has a recognizable style.

In addition, we also ask our expert annotators (i) whether the text has been written by a machine or a human writer, (ii) if their opinion would match other experts' opinion; and (iii) if their opinion would match the opinion of general readers.

\textbf{Evaluators.} We recruited six literary experts, all critics or university scholars. These experts were different from those who developed the rubric. Three of them evaluated the 60 Spanish texts written by Pron, and the 60 Spanish texts written by GPT-4. The other three were bilingual and experts in English Literature, and evaluated the 60 Spanish texts by Pron and the 60 English texts written by GPT-4. More information on the evaluators can be found in Appendix \ref{appx:assesors}.

\section{Results and Discussion}
In this section, we present the main findings of our study. Each subsection provides a detailed analysis of the expert assessment annotations in order to answer our research questions.
\subsection{RQ1: Can the current state of generative AI compare to a prestigious author in creative writing tasks?}

\begin{figure*}[htbp]
    \centering
    \subfloat[GPT-4 Turbo in English]{%
        \includegraphics[width=0.47\textwidth]{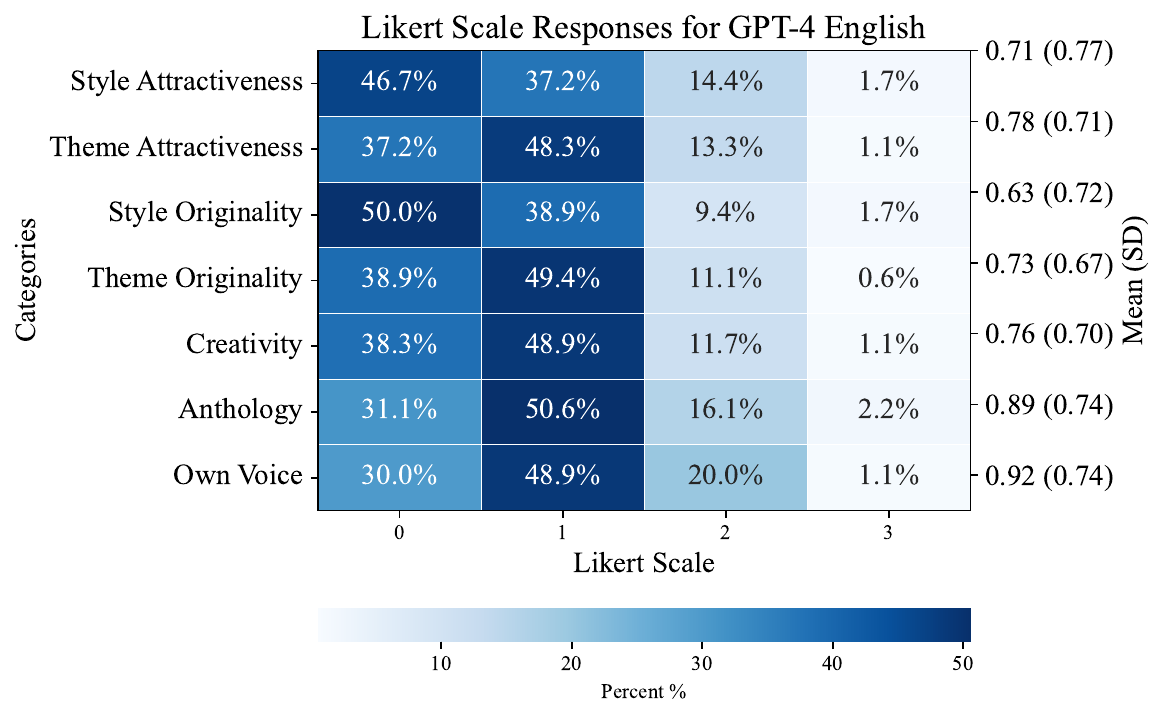}
    }
    \hfill
    \subfloat[GPT-4 Turbo in Spanish]{%
        \includegraphics[width=0.47\textwidth]{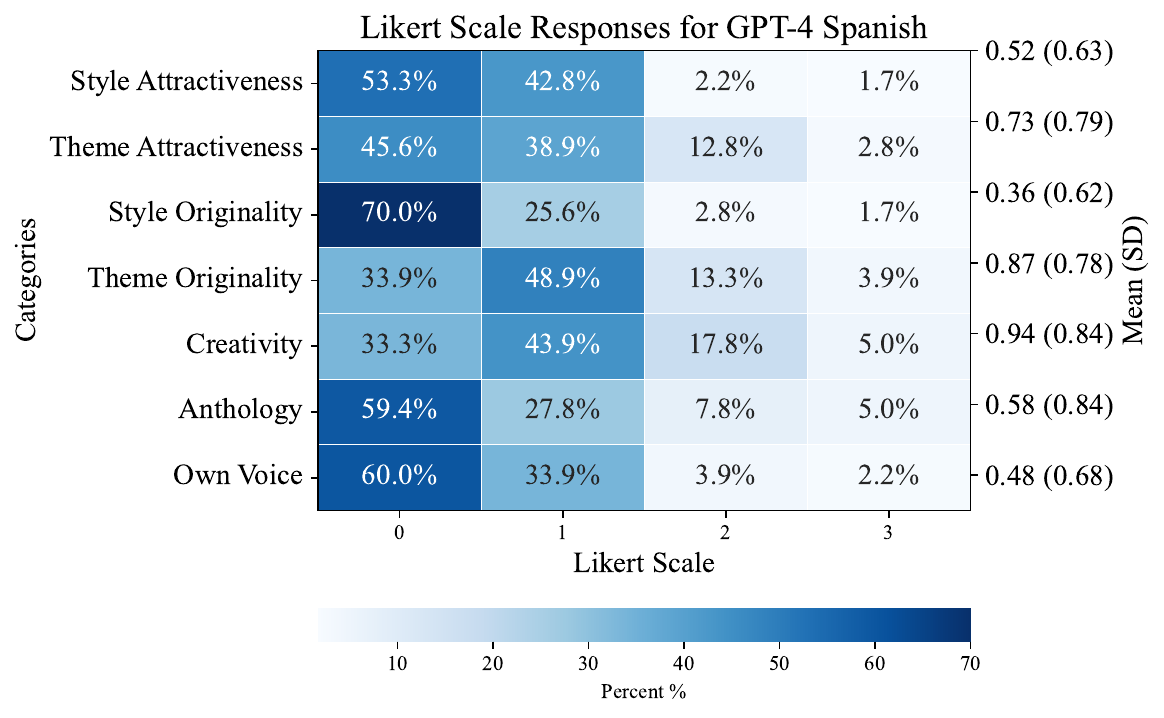}
    }
    \hfill
    \subfloat[Patricio Pron]{%
        \includegraphics[width=0.47\textwidth]{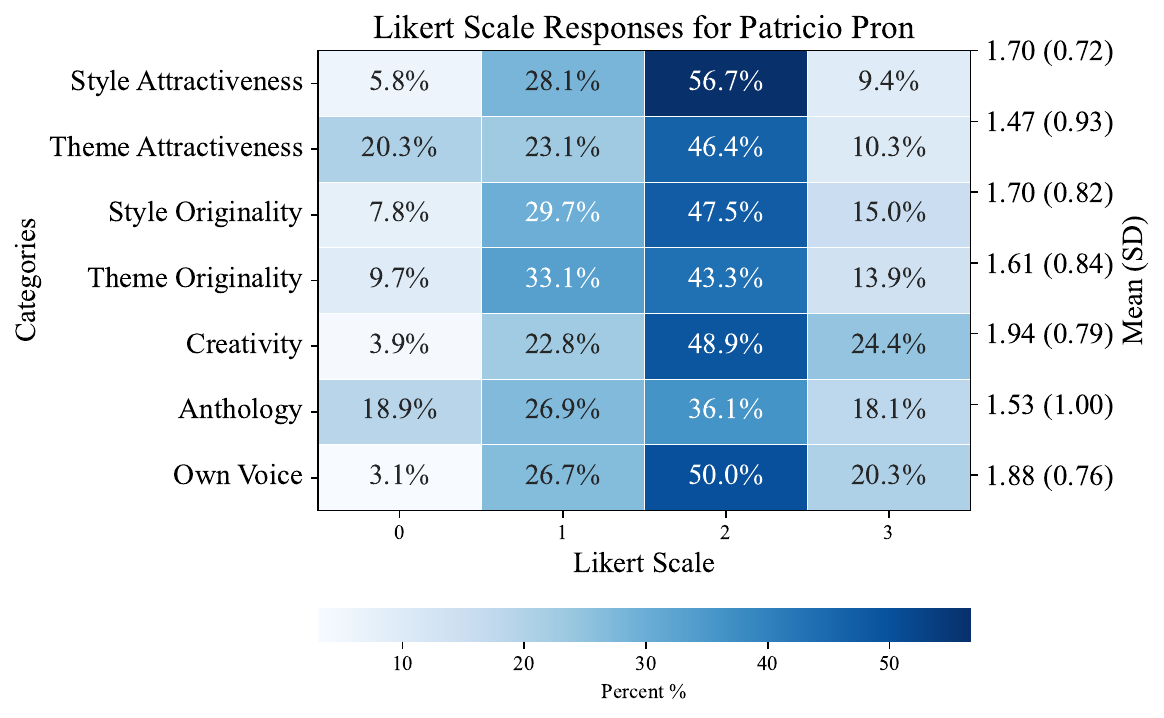}
    }
    \caption{Summary of expert assessments for each writer}
    \label{fig:tres-subfiguras}
\end{figure*}

Figure \ref{fig:tres-subfiguras} summarizes the scores given by the experts to GPT-4 (English and Spanish are reported separately) and Patricio Pron, showing the percentage of assessments in each of the 0-3 scores and also the mean and standard deviation for each of the quality dimensions in the rubric\footnote{Although computing means is not advised with likert scores, considering that the numeric scores follow a ratio scale and were provided to the assessors in the rubric, we report them as complementary descriptive figures.}. 

Overall, the assessments are remarkably lower for GPT-4 across all quality dimensions, in both languages. In all dimensions, GPT-4 receives predominantly scores of 0/1, while Pron receives mostly 2/3. Pron writes with more original and attractive style and theme, is more creative, his texts are more suitable to be included in an anthology, and assessors believe that he has its own voice compared to GPT-4. Pron's creativity score is particularly high (average of 1.94), which roughly doubles GPT-4 creativity scores in English and Spanish. On the other hand, GPT-4 scores are particularly low in style originality (0.36 in Spanish and 0.63 in English), where Pron gets 1.70 (more than double). 

Note that a direct comparison between GPT-4 scores in English and Spanish is not advisable, as they come from different groups of assessors. In Section 4.3 we discuss language differences. 

Overall, GPT-4's concentration of evaluations in the lower scoring brackets and its substantially lower mean scores provide evidence that, while the LLM can generate coherent and structurally sound text, it still lacks the depth, originality, and stylistic nuance that characterize a top fiction writer such as Patricio Pron. Pron's higher scores and favorable evaluations across all criteria underscore his ability to produce engaging, original, and creatively rich content.

This data suggests that the answer to our main research question —--\textit{Can the current state of generative AI compare to a prestigious author in creative writing tasks?}--- is currently no. GPT-4 is not yet on par with top fiction writers, and the difference is so large that it is unlikely that some of its current peers --Claude, Gemini, Llama-- could obtain significantly different results. Despite  advancements in language modeling, and the ability of LLMs to produce grammatically correct and contextually relevant text, there remains a significant gap in terms of originality, stylistic attractiveness, and the conveyance of a unique authorial voice. 

Note, however, that although GPT-4 does not resist the comparison in average, there are a few cases where its texts receive the highest possible scores from literature experts, which is still remarkable. For instance, in the question about creativity (which encompasses some of the more fined-grained questions), its texts receive the highest score from an evaluator in eleven occasions (though never twice for the same text). The fact that GPT-4 can occasionally receive the highest possible evaluation of its creativity from experts is a positive note on its potential for fiction writing, and a testimony of the advance in the field over the last few years. 


\subsection{RQ2: What is the role of the prompt (the proposed title) in the creativity of the result?}

\begin{figure*}[t]
  \centering
  \begin{subfigure}[b]{0.49\linewidth}
    \includegraphics[width=\linewidth]{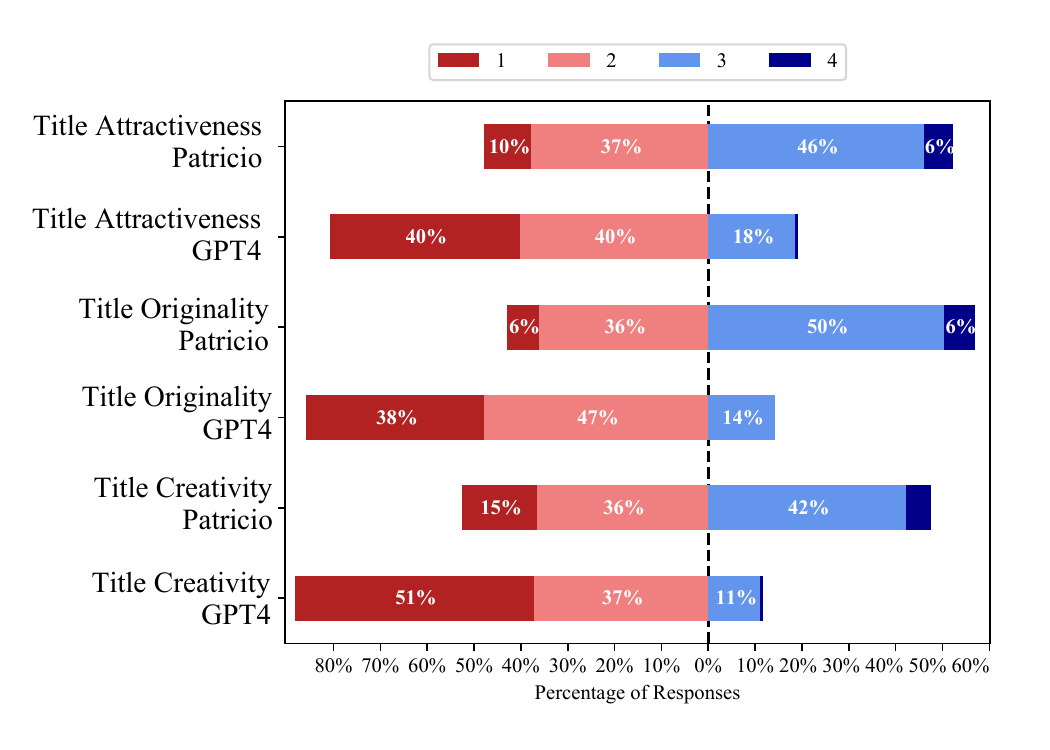}
    \caption{Percentage of Likert scores (0-4) received by Pron and GPT-4 titles on all quality dimensions.}
    \label{fig:titles_comparison}
  \end{subfigure}
  \hfill
  \begin{subfigure}[b]{0.49\linewidth}
    \includegraphics[width=\linewidth]{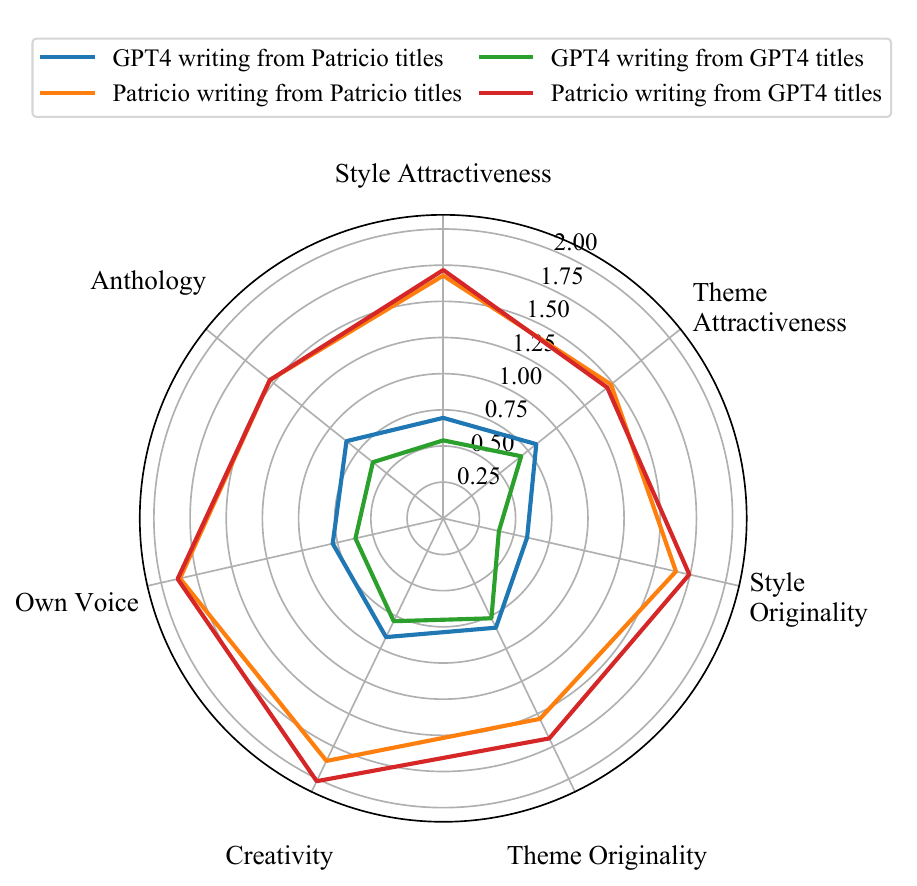}
    \caption{Comparison of the impact of using Pron's titles versus GPT-4 titles on the text quality. Values are averages of the likert scores received.}
    \label{fig:radar_plot}
  \end{subfigure}
  \caption{Influence of the prompt in the creativity of texts: quality of Pron vs GPT-4 titles on the left, and quality of the texts produced with each type of title on the right.}
  \label{fig:prompt}
\end{figure*}

Figure \ref{fig:prompt} shows the experimental results for this research question. The leftmost figure shows that Pron's titles receive significantly higher scores in originality, attractiveness and creativity than its GPT-4 peers. Then, do better titles lead to better texts?


In Figure \ref{fig:radar_plot} (right)  we can see the effect of both sets of titles in the texts written for them. The figure shows a radar chart with average likert scores for five quality dimensions. Remarkably, GPT-4 receives better scores in all quality dimensions when the titles have been provided by Pron. Differences are particularly high in style originality (+57\%), style attractiveness (+30\%), suitability for an anthology (+45\%), and author having its own voice (+30\%). A mere title provided by a creative writer can induce the LLM to produce texts with a better creative style. In contrast, the quality of Pron texts seems to be mostly independent of the provenance of the title and, for some quality dimensions, the (less creative) GPT-4 titles seem to be a challenge that Pron resolves with even higher average scores: theme originality is 10\% better with GPT-4 titles, style originality is 6\% better, and creativity is 9\% better. We asked Pron about this and he replied that "I did not like GPT-4 titles at all, so I tried to take them in completely different directions". 

In order to find out if the differences are statistically significant, we used the Mann-Whitney U test \cite{mcknight2010mann}, a non-parametric test that is ideal for comparing differences between two not-paired independent groups when the data does not necessarily meet the assumptions required for parametric tests. According to this test, GPT-4 improved scores when using Pron's titles are statistically significant for \textit{Style Originality} ($p=0.01$), suitability for an \textit{Anthology} ($p=0.01$), \textit{Theme Attractiveness} ($p=0.04$) and \textit{Own Voice} ($p=0.03$). The other two dimensions receive $p=0.06$ (\textit{Style Attractiveness}), and $p=0.15$ (\textit{Creativity}). 


Overall, these results support the hypothesis that the creative request is a crucial factor in the behavior of LLMs when producing creative text writing, to the point that a mere prompt is perhaps enough to talk about co-authorship. The results also suggest that human-machine collaboration in the creative writing arena has more potential than completely autonomous LLM writing.

Our human writer, on the other hand, seems capable to cope with worse titles, and even use them as a motivating creative constraint that results in even (slightly) better texts. 

\subsection{RQ3: Are models more creative in English than in Spanish?}

\begin{figure}[t]
\centering
\includegraphics[width=\linewidth]{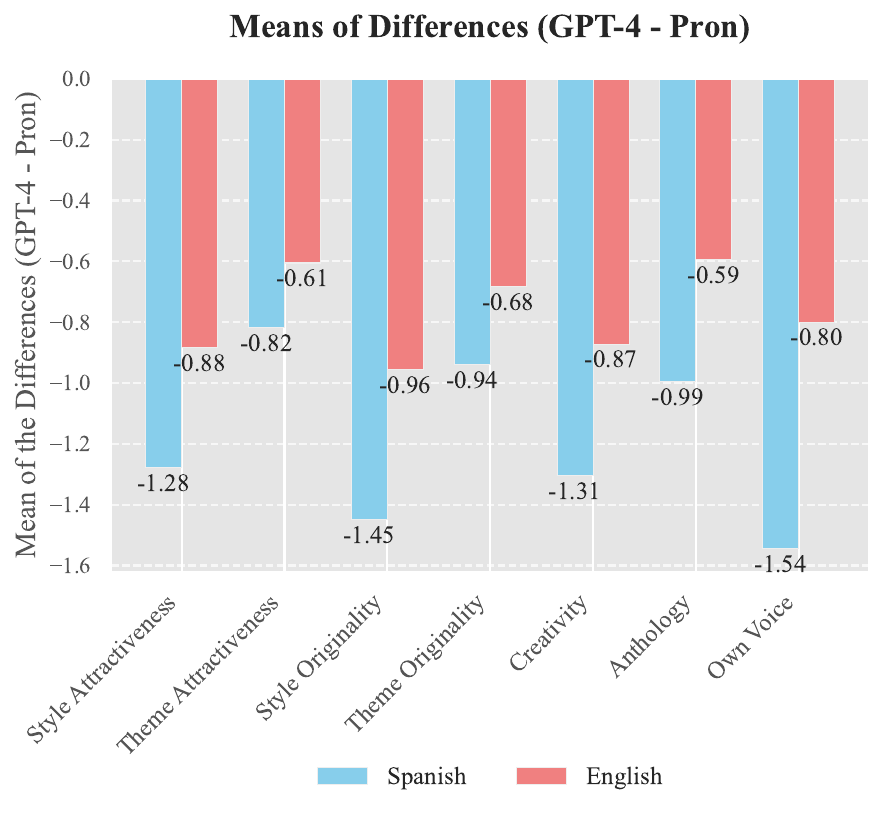} \hfill
  \caption {GPT-4 performance differences with Pron in English and Spanish (a negative score indicates that GPT-4 scores lower than Pron)}
    \label{fig:EN-ES-comparison}

\end{figure}


In order to analyze the performance in both languages,  we calculated the intra-individual score differences between Pron and GPT-4 for each evaluator-title pair. By focusing on these differences, we aimed to minimize variance caused by individual evaluator biases. Since Pron's stories were identical for both evaluator groups (English and Spanish), we were able to directly compare the gap between GPT-4's performance in each language. The results, shown in Figure~\ref{fig:EN-ES-comparison}, indicate that GPT-4's gap with Pron is consistently larger in Spanish than in English, with the most significant disparity observed in \textit{Own Voice}, where the gap nearly doubles from -0.8 in English to -1.54 in Spanish. 

To assess the statistical significance of the mean differences observed, we performed paired statistical tests for each of the quality dimensions. For attributes where differences were normally distributed (according to a Shapiro-Wilk test \cite{shapiro1965analysis}), we used a paired t-test, and for the rest of attributes we used a non-parametric Wilcoxon signed-ranked test \cite{wilcoxon1992individual}. Finally, we used the Bonferroni correction \cite{bonferroni1936teoria} to adjust the significance threshold based on the number of tests conducted. The outcome of the statistical analysis is summarized in Table~\ref{tab:stat-test}: GPT-4 is significantly better in English for all quality dimensions, except the two related to theme (theme attractiveness and theme originality). Note that these two dimensions are the ones less related to language itself, which reinforces the conclusion that the differences observed are related to differences in linguistic competence in both languages.

\begin{table*}[htbp]
\footnotesize
\centering
\caption{Paired statistical test results for attribute differences between English and Spanish} \label{tab:stat-test}
\label{tab:stat_results}
\begin{tabular}{llcccc}
\toprule
\textbf{Attribute} & \textbf{Test Used} & \textbf{Statistic} & \textbf{p-value} & \textbf{Corrected p-value} & \textbf{Significant} \\ 
\midrule
Style Attractiveness & Paired t-test & -4.215 & $8.68 \times 10^{-5}$ & $6.08 \times 10^{-4}$ & Yes \\
Theme Attractiveness & Paired t-test & -1.689 & $9.66 \times 10^{-2}$ & $6.76 \times 10^{-1}$ & No \\
Style Originality & Paired t-test & -5.757 & $3.26 \times 10^{-7}$ & $2.28 \times 10^{-6}$ & Yes \\
Theme Originality & Paired t-test & -2.429 & $1.82 \times 10^{-2}$ & $1.28 \times 10^{-1}$ & No \\
Creativity & Wilcoxon signed-rank & 231.000 & $4.91 \times 10^{-5}$ & $3.44 \times 10^{-4}$ & Yes \\
Anthology Potential & Paired t-test & -3.445 & $1.06 \times 10^{-3}$ & $7.41 \times 10^{-3}$ & Yes \\
Own Voice & Wilcoxon signed-rank & 126.500 & $2.53 \times 10^{-8}$ & $1.77 \times 10^{-7}$ & Yes \\ 
\bottomrule
\end{tabular}
\end{table*}

\subsection{RQ4: Does GPT-4 have a recognizable style for a literature expert when generating creative text?} 

To answer this question, we want to measure if the ability to detect LLMs authored text improves along the evaluation process, i.e., if the experts learn about the traits of GPT-4 writing vs Pron's writing by reading its texts (even if the expert is not informed about authorship). 

\begin{figure}[t]
\centering
\includegraphics[width=\linewidth]{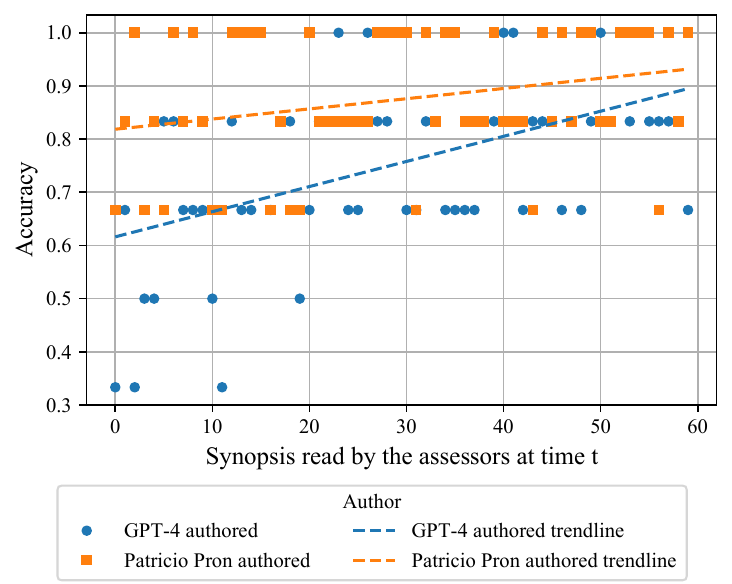} \hfill
  \caption {Evolution of accuracy detecting AI-Generated and Human-Written Texts over time.}
    \label{fig:accuracy_time}

\end{figure}

Figure \ref{fig:accuracy_time} displays the accuracy of identifying texts written by GPT-4 and by human writers over the full sequence of 60 texts. The x-axis represents the order of the texts from the first to the sixtieth, while the y-axis shows averaged evaluators' accuracy.

The two main lines represent the evaluators' accuracy trend in identifying AI-generated texts (blue line) and human-written texts (orange line). Note that the accuracy in detecting human-written texts is consistently high (with a slight increase with time). In contrast, the accuracy in detecting AI-generated texts is more variable, and shows a higher learning slope over time. This indicates that evaluators learn to recognize writing patters in GPT-4 as they gain more experience. 

Overall, these results suggest that, in the absence of stylistic directions, the creative writing style of LLMs may have recognizable traits. 

\textbf{RQ5: Is Boden's definition of creativity operational when assessing creative text writing?}

\begin{table}
  \centering
  \scriptsize
    \begin{tabular}{lrrr}
    \hline
     & \textbf{Attractiveness} & \textbf{Originality} & \textbf{Creativity} \\
    \hline
    \textbf{Attractiveness} & 1.0 &  & \\
    \textbf{Originality} & 0.78 & 1.0 &  \\
    \textbf{Creativity} & 0.72 & 0.73 & 1.0 \\
    \hline
    \end{tabular}
  \caption{Spearman correlation for the dimensions of attractiveness, originality, and creativity.}
  \label{tab:spearman}
\end{table}

We assessed creativity by mapping Boden's dimensions into attractiveness and originality of both theme and style, and we also asked assessors to evaluate creativity as a whole. Do Boden's dimensions correlate with creativity assessments? 
Table \ref{tab:spearman} shows Spearman correlations between creativity, attractiveness and originality. All variables are correlated with values above 0.7, which is a strong positive signal. Note, however, that the relation between originality and attractiveness is higher (0.78) than the relation of each of the components with creativity (0.73 and 0.72), which suggests that the relationship is nuanced and may depend on each assessor's take of what is creativity. 

\begin{figure}[t]
\centering
  \includegraphics[width=.7\linewidth]{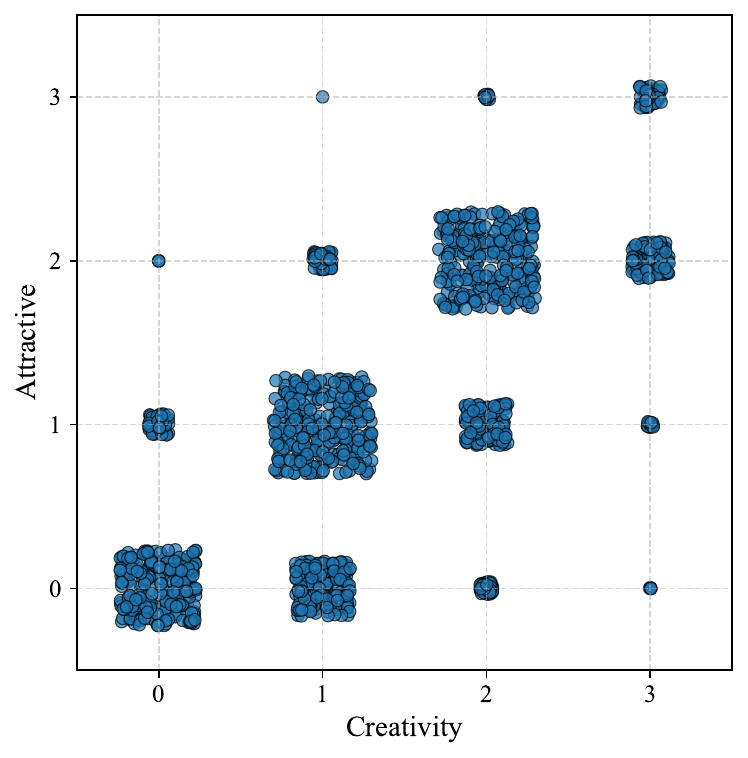} \hfill
  \includegraphics[width=.7\linewidth]{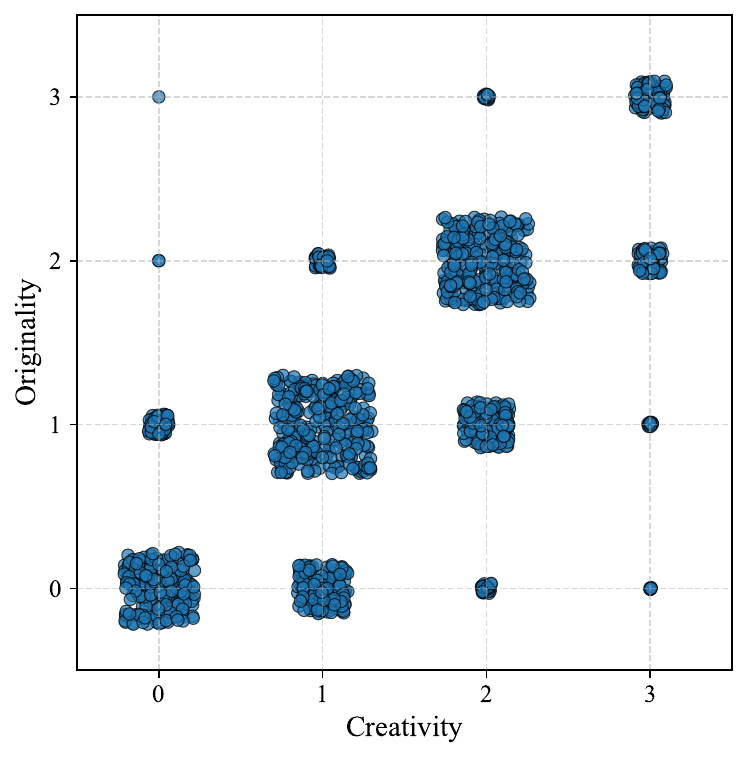}
  \caption{Correlation plots for creativity versus attractiveness and originality.}
  \label{fig:corr_plots}
\end{figure}

Figure \ref{fig:corr_plots} confirms these correlations visually. Each dot represents an expert assessment for a given text. It shows a much higher density of points along the diagonal, suggesting a positive correlation between these dimensions, both in terms of style and theme. In addition, it seems that attractiveness and originality are upper bounds for creativity, because the zones below the diagonal are more populated than the zones above the diagonal in both graphs. 

To robustly test this, we applied mixed-effects models \citep{bates2014fitting}, accounting for variability in titles and evaluators. The model we fit is: \textit{\(\text{Creativity} = \beta_0 + \beta_1 \text{Style\_Attractiveness} + \beta_2 \text{Theme\_Attractiveness} + \beta_3 \text{Style\_Originality} + \beta_4 \text{Theme\_Originality} + u_{\text{title}} + v_{\text{username}} + \epsilon\).} We obtain significant contributions from all predictors ($p < 0.001$). Results show that the REML criterion at convergence is $991.7$, with scaled residuals between $-4.47$ and $3.59$. Variance components are $0.006$ for titles, $0.079$ for evaluators, and $0.21$ for residuals.  The fixed effects are: \textit{Style Attractiveness} ($estimate = 0.18$, \( p < 0.001 \)), \textit{Theme Attractiveness} ($est. = 0.15$, \( p < 0.001 \)), \textit{Style Originality} ($est. = 0.33$, \( p < 0.001 \)), and \textit{Theme Originality} ($est. = 0.33020$, \( p < 0.001 \)).

The intercept is estimated at $0.25$ with marginal significance ($p = 0.08$). Correlations between fixed effects are low, the highest being $-0.712$ between \textit{Style Attractiveness} and \textit{Style Originality}. This analysis shows that both attractiveness and originality contribute significantly to creativity, with originality having a slightly stronger impact.

In summary, all predictors have an impact on creativity, with originality playing a more prominent role; and high residual variability suggests other factors influencing creativity which are not capture by the model.

\section{Conclusions}


Our results indicate that GPT-4 Turbo, despite its impressive writing capabilities, still falls short of matching the skills of a world-class novelist. Texts generated by GPT-4 are consistently rated lower in all quality dimensions in our study: attractiveness and originality of both style and theme, and overall creativity, among others. Comparing with previous results, this indicates that it is much easier to match the average performance of human writers than to actually match the best ones: LLMs still lack the nuanced depth, originality and intent characteristic of a top novelist such as Patricio Pron. 

Also, our study highlights the significant role of prompts in creative text writing: titles provided by Pron resulted in GPT-4 texts which are significantly more creative and original than the ones written for its own titles. Even the simplest prompting (short titles in our case) should be considered co-authorship, as it has a profound influence on the results.  
    
We also found that GPT-4's performance in generating creative texts was significantly better in English than in Spanish, in spite of being also a resource-rich language. This discrepancy is likely due to the model being trained on a larger corpus of English text, reflecting a bias towards English in the available training data. The results underscore the need for more balanced and comprehensive training datasets to enhance the multilingual creative writing capabilities of AI systems.
    
Our expert evaluators were able to identify AI-generated texts with increasing accuracy over time, suggesting that GPT-4 has a recognizable style that becomes more apparent as evaluators gain experience with its outputs. This indicates that despite its ability to mimic human writing, GPT-4's generated text retains a certain uniformity that can be detected by expert readers. 
    
Finally, our study successfully applied Boden's approach to creativity (as a combination of novelty, surprise, and value) to create a rubric that serves to evaluate creative writing texts, either human or machine-generated. A statistical analysis of the 7,200 manual assessments collected shows that both attractiveness (value) and originality (novelty and surprise) significantly contribute to the perceived creativity of texts. This validates the use of Boden's dimensions in evaluating the creative outputs of AI systems.

 It is reasonable to conclude that there are inherent limitations in current LLMs. LLMs rely on pattern extraction from large corpora of text, which allows them to generate text that is contextually appropriate and often mimics the stylistic nuances of human writing. However, this approach can also lead to the generation of content that tends to conform to common patterns and clichés, which may be enough when compared to average professional writers, but lacks the originality and intent found in the best human writers.

Probably, a key limitation of LLMs is their tendency to approximate meaning through probability. While human writers can produce low-probability text that carries deep meaning and creativity, LLMs tend to generate content that aligns with the most likely patterns observed in their training data. This probabilistic approach can result in outputs that are high in coherence and fluency but low in innovative thinking and originality. As LLMs are refined and improved, they are likely to become more adept at solving objective tasks. However, their creative output may remain constrained by a tendency to replicate familiar patterns, leading to a literature filled with clichés.

Overall, our study suggests that while LLMs can be valuable tools for generating text and assisting with various writing tasks, they are not yet capable of fully replicating the creative process of top human writers, who often produce work that is not only meaningful but also surpass conventional expectations. For now, we will have to wait until a duel between top human and machine writers is actually disputed.

All experimental data and texts are available for reproducibility purposes at \url{https://github.com/grmarco/pron-vs-prompt}.

\section*{Limitations}

These are the main limitations of our work:

\begin{itemize}
    \item \textbf{Prompt design and influence in the results}: Careful prompt engineering would imply a de-facto collaboration between man and machine; therefore, to avoid contamination we decided not to fine-tune our prompts in any way, and simply provide similar instructions to our human writer and to GPT-4, without further fine-tuning. This means that there might be alternative prompts that result in better GPT-4 texts that we have not explored. 
    \item \textbf{Limited scope of our creative writing task}: The study focused on a specific creative writing task: writing short synopsis for imaginary films with a given title. Creativity writing encompasses a broader range of tasks which were not evaluated. Consequently, our findings may not be generalizable to other forms of creative expression where different skills and qualities are required. Also, for larger texts (such as a novel), internal coherence may be a challenge for LLMs, which is not an issue in our experimental setup. 
    
    \item \textbf{Scope of language and cultural contexts}: The study only considered texts in English and Spanish, limiting the scope of our findings. Creativity is deeply influenced by cultural context, and our study does not account for the vast diversity of linguistic and cultural nuances across other languages. In any case, we would expect to find an even larger gap between GPT-4 and top human writers in other languages with less online resources.
    
    \item \textbf{Focus on a Single AI model}: While GPT-4 is a state-of-the-art language model, it represents only one approach to AI text generation. Other models, possibly with different architectures or training paradigms, might exhibit different strengths and weaknesses in creative tasks. Our study does not account for these variations, potentially limiting the applicability of our findings to a broader range of AI systems. 
    
    \item \textbf{Multilingual design}: In order to avoid undesired translation effects, Pron texts were kept in its original language (Spanish) for all evaluators. Our bilingual experts (all scholars in English literature with bilingual language skills) evaluated GPT-4 texts in English together with Pron texts in Spanish. Although results are consistent with the Spanish evaluation, there might be undetected effects of language in the comparative evaluation of GPT-4 english texts. In particular, the decision of authorship might be influenced by the fact that all English texts had been written by GPT-4, which was an easy to spot signal. In average they were not, however, better authorship predictors than their monolingual counterparts. 

    \item \textbf{Only expert assessments}: There are always two types of veredict for a creative text: the opinion of the experts (critics and scholars), and the reception of the audience (the readers). Both are relevant and not always correlate with each other. We have only collected experts' assessments, so the question of whether the audience would perceive a similar gap between Pron and GPT-4 texts remains open. 
    
\end{itemize}

In view of these limitations, future research should consider: 

\begin{itemize}
    \item Expanding the scope of creative tasks and considering man-machine co-authoring processes, including prompt engineering techniques.
    \item Incorporating readers (the audience, rather than the critics) as evaluators to capture a broader notion of value in the experimentation.
    \item Exploring other models and architectures to identify different approaches to enhance the creativity of AI systems.
\end{itemize}


\section*{Acknowledgements}
This work has been financed by the European
Union (NextGenerationEU funds) through the “Plan
de Recuperación, Transformación y Resiliencia”, by
the Ministry of Economic Affairs and Digital Transformation and by UNED University. However,
the points of view and opinions expressed in this
document are solely those of the authors and do
not necessarily reflect those of the European Union
or European Commission. Neither the European
Union nor the European Commission can be considered responsible for them.

Guillermo Marco's work was funded by Spanish government Ph.D. research grant ({\it Ministerio de Universidades}) FPU20/07321 and a scholarship of the Madrid City Council for the Residencia de Estudiantes (Course 2023-2024).

\bibliography{custom}

\appendix

\section{Rubric for the Evaluation}
\label{apxx:rubric}
The form used in this research is structured in three blocks, each designed to assess different aspects related to creativity: the dimensions of creativity, authorship and the process of assessing the synopsis. The evaluators were constantly provided with the title and synopsis of each text to ensure that their assessments were accurate and consistent with the text being evaluated.

Each dimension is assessed using a Likert scale \cite{matell1971}, whose scale from 0 to 3 was accompanied by qualitative descriptions for each value.

In terms of attractiveness, three aspects are asked for assessment: title, style and theme. The questions are as follows:

\begin{enumerate}
    \item Rate the attractiveness of the following aspects of the text, understanding it as a literary object:
    \begin{itemize}
        \item Do you find the title attractive? Does it catch your attention and make you interested in the synopsis?
        \begin{itemize}
            \item[0] It does not catch your attention, nor does it generate any interest in the story.
            \item[1] Hardly catches your attention, but does arouse mild interest.
            \item[2] It is thought-provoking and arouses curiosity.
            \item[3] It is captivating, generates a lot of expectations about the text.
        \end{itemize}
        \item Do you find the style appealing and does it make you enjoy reading the synopsis?
        \begin{itemize}
            \item[0] The style lacks appeal and even discourages reading.
            \item[1] The style has a slight appeal but does not particularly stand out.
            \item[2] The style is appealing and contributes to reading enjoyment.
            \item[3] The style is engaging in its own right, creating a memorable reading experience.
        \end{itemize}
        \item Do you find the story and characters engaging and do they capture your interest in the subject matter itself?
        \begin{itemize}
            \item[0] The story and characters lack appeal, with no elements that stand out or add value.
            \item[1] The story and characters have some appeal, but lack attention-grabbing elements.
            \item[2] The story and characters are quite appealing and attention-grabbing.
            \item[3] The story and characters are very appealing, and are fully attention-grabbing.
        \end{itemize}
    \end{itemize}
    \item Assess the originality of the following aspects of the text. We understand that a text is original if it surprises you (if it is something you did not expect, if it is unfamiliar or reminds you of things you have read before) regardless of whether you found it attractive or not.
    \begin{itemize}
        \item Does the title seem original to you?
        \begin{itemize}
            \item[0] The title is very predictable, it is a pure cliché.
            \item[1] The title is quite predictable and resorts to commonplaces.
            \item[2] The title is quite original and avoids commonplaces.
            \item[3] The title is unique and surprising, unlike anything I have seen before.
        \end{itemize}
        \item Is the style of the text original and surprising?
        \begin{itemize}
            \item[0] The style is formulaic and completely conventional.
            \item[1] The style has some original elements, but is predominantly conventional.
            \item[2] The style is relatively original, and it is not easy to relate it to other writers.
            \item[3] The style is highly original, and reveals a voice of the writer.
        \end{itemize}
        \item Is the plot of the text original and innovative?
        \begin{itemize}
            \item[0] The plot is completely conventional and resorts to widely explored ideas or clichés without bringing in new elements.
            \item[1] The plot is fairly conventional, although there are marginal attempts to introduce original elements.
            \item[2] The plot is quite original and brings in innovative elements.
            \item[3] The plot is highly original and innovative.
        \end{itemize}
    \end{itemize}
    \item Bearing in mind that the title was the starting point of the literary exercise, do you think the writer has used it well?
    \begin{itemize}
        \item I don't see any relationship between the title and the synopsis.
        \item The relationship between the title and the synopsis is almost irrelevant.
        \item There is some relationship between the title and the synopsis, though not obvious.
        \item There is a lot of relationship between the title and the synopsis and, in fact, the development of the text is predictable from the title.
        \item There is a strong link between the title and the synopsis and, moreover, the link is surprising, the way the title is developed is original.
    \end{itemize}
    \item Please rate the creativity of (the title and synopsis):
    \begin{itemize}
        \item[0] Not creative at all.
        \item[1] Slightly creative.
        \item[2] Quite creative.
        \item[3] Very creative.
    \end{itemize}
    
\end{enumerate}

In the second block of the questionnaire, the author's assessment and perspective on the creative process is collected. This block helps to identify possible biases or influences of the evaluators in the assessment of creativity. The questions are:

\begin{enumerate}
    \setcounter{enumi}{4}
    \item Assess authorship:
    \begin{itemize}
        \item Who do you think wrote this title? Who do you think wrote the synopsis?
        \begin{itemize}
            \item An amateur writer.
            \item An established writer.
            \item An artificial intelligence.
        \end{itemize}
    \end{itemize}
    \item The exercise of inventing a synopsis for an imaginary film can be seen as a kind of literary genre. Imagine that such a genre exists:
    \begin{itemize}
        \item Do you think you would select this text in an anthology of this genre?
        \item Do you think your assessment of the text would match that of most readers?
        \item Do you think your assessment of the text would match that of most literary critics?
        \item Judging by this text alone, do you think it is likely that the author has a recognisable style, i.e. a voice of his own?
    \end{itemize}
\end{enumerate}

Finally, because of the interest of the qualitative aspects for this research, each of the three blocks ends with an optional question to comment on the answer. In the same way, a final questionnaire was carried out so that the evaluators could comment on what they thought of the evaluation process, as well as the questions in the questionnaire.

\section{Academic and Professional Profile of the Review Panel}
\label{appx:assesors}

The review panel consists of six members with expertise in writing literature and translation. Three male reviewers aged 45-55 are teachers with diverse academic backgrounds: one holds degrees in Hispanic Philology and Comparative Literature while the other two have Master's Degree in Narrative. Two female reviewers, also aged 45-55, are lecturers in Translation, both holding PhDs in Translation. Finally, one male reviewer, aged 25-35, is a professional translator with a degree in Translation and Interpretation. The evaluators were paid 25 euros/hour. Table \ref{tab:assesors} summarizes the information on the evaluators.

\begin{table*}[htbp]
	\centering
	\scriptsize
	\begin{tabular}{|l|l|p{4cm}|l|l|p{4cm}|}
		\hline
		\textbf{Identifier} & \textbf{Language of review} & \textbf{Profession} & \textbf{Age range} & \textbf{Gender} & \textbf{Educational Background} \\
		\hline
		Reviewer 1 & Spanish & Teacher of Language \newline and Literature, Literary Critic & 45-55 & Male & Hispanic Philology, \newline Theory and Comparative Literature \\
		\hline
		Reviewer 2 & Spanish & Teacher of Creative \newline Writing, Writer & 45-55 & Male & Chemistry, \newline Master's Degree in Narrative \\
		\hline
		Reviewer 3 & Spanish & Teacher of Creative \newline Writing, Writer & 45-55 & Male & Computer Science, \newline Master's Degree in Narrative \\
		\hline
		Reviewer 4 & English, Spanish & Lecturer in Translation, \newline Translator & 45-55 & Female & Translation and Interpretation, \newline PhD in Translation \\
		\hline
		Reviewer 5 & English, Spanish & Lecturer in \newline Translation & 45-55 & Female & English Philology, \newline Postgraduate degrees in Translation \newline and Comparative Literature, \newline PhD in Translation \\
		\hline
		Reviewer 6 & English, Spanish & Translator & 25-35 & Male & Translation and Interpretation \\
		\hline
	\end{tabular}
	\caption{Academic and Professional Profiles of the Review Panel}
	\label{tab:assesors}
\end{table*}

\section{Prompts and Materials}
\label{appx:materials}
In this appendix, we report the prompts and several synopses that we used in the experiment.
\subsection*{Titles proposed by Patricio Pron}

\begin{enumerate}
\small
    \item After all I almost did for you
    \item All love songs are sad songs
    \item Another episode in the Class Struggle
    \item Don't tell mom
    \item Eclipse in the botanical garden
    \item Edith loves him (we'll come back to this)
    \item Every picture from when we were young
    \item Future ghosts
    \item I have no fear because I have nothing
    \item I keep trying to forget your promise
    \item Lindsay Hilton visits Paris
    \item Mental illness three days a week
    \item Monsters live here
    \item Paradise can't be seen from here
    \item Pick a card, any card. No, not that one! Another!
    \item Rise and fall of R. S. Turtleneck, children's author
    \item Silks from Bursa, tiles from Kütahya
    \item Spanish Youth, keep trying
    \item The day after Groundhog day
    \item The delights of the garden of delights
    \item The last journey of Santiago Calatrava
    \item The last laugh of that year
    \item The Lego woman
    \item The national red button
    \item The nightmares of the invisible man
    \item The nocturnal emissions
    \item The tied cow
    \item Two cops stand between us
    \item When you are at the top you can't fall any lower
    \item Who killed Patricio Pron?
\end{enumerate}

\section*{Titles proposed by GPT-4}

\begin{enumerate}
\small
    \item Among clouds and mirages
    \item Between the lines of fate
    \item Beyond the broken horizon
    \item Bits of reality
    \item Echoes of a lost dream
    \item Echoes of the future
    \item Fragments of an invisible yesterday
    \item Parallel paths
    \item Reflections of another world
    \item Shadows in the mist
    \item Song of the captive moon
    \item Sparks in the dark
    \item The awakening of the aurora
    \item The crystal labyrinth
    \item The echo of silenced voices
    \item The forgotten melody
    \item The garden of withered dreams
    \item The inverted city
    \item The journey of the dawn
    \item The last flight of the butterfly
    \item The last night on Earth
    \item The mosaic of time
    \item The painter of memories
    \item The shadows of time
    \item The whisper of the cosmos
    \item The wind in the moorlands
    \item Traces in the sea of sand
    \item Twilight of the titans
    \item Under the copper sky
    \item Whispers from the eternal city
\end{enumerate}

\begin{table*}[htbp]
	\centering
	\scriptsize
	\begin{tabular}{|p{7cm}|p{7cm}|}
		\toprule
		\textbf{English Prompt} & \textbf{Spanish Prompt} \\
		\midrule
		\textbf{System Prompt}: \newline We are going to do an experiment in which we are going to compare \newline your creative writing skills with those of a prestigious novelist, \newline Patricio Pron. The task is to generate synopses for movie titles \newline that do not exist. The synopses must be creative and appealing to both \newline critics and the general audience, and must have literary value in \newline and of themselves. \newline Here are some details of the novelist you will be competing with: \newline Patricio Pron (Rosario, December 9, 1975) is a writer and literary critic. \newline Granta magazine selected him in 2010 as one of the 22 best young \newline writers in Spanish. He won the twenty-second Alfaguara Novel \newline Prize in 2019 for his work Mañana tendremos otros nombres.
		&
		\textbf{System Prompt}: \newline Vamos a hacer un experimento en el que vamos a comparar tus \newline habilidades de escritura creativa con las de un novelista prestigioso, \newline Patricio Pron. La tarea consiste en generar sinopsis para títulos de \newline películas que no existen. Las sinopsis deben ser creativas y atractivas \newline tanto para los críticos como para el público en general, y deben tener \newline valor literario por sí mismas. \newline \newline Aquí tienes algunos detalles del novelista con el que competirás: \newline Patricio Pron (Rosario, 9 de diciembre de 1975) es escritor y crítico \newline literario. La revista Granta lo seleccionó en 2010 como uno de los 22 \newline mejores jóvenes escritores en español. Ganó el vigésimo segundo \newline Premio Alfaguara de Novela en 2019 por su obra Mañana tendremos \newline otros nombres. \\
		\midrule
		\textbf{User Prompt}: \newline The proposed title is: "{title}". Please write a synopsis of about \newline 600 words for that title that meets the above specifications.
		&
		\textbf{User Prompt}: \newline El título propuesto es: "{titulo}". Por favor, escribe una sinopsis de \newline aproximadamente 600 palabras para ese título que cumpla con las \newline especificaciones mencionadas anteriormente. \\
		\bottomrule
	\end{tabular}
	\caption{System and User Prompts}
\end{table*}

\subsection*{Example Synopses}

\textbf{1. Best synopsis by Pron (average score 2,29)}

\noindent\textit{Title: Fragmentos de un ayer invisible}


{\small
Walt Disney es «devuelto a la vida» por sus descendientes en el momento en que éstos, que tienen una relación difícil y grandes diferencias respecto al futuro de la compañía, acuerdan ponerla en manos de su creador. Disney regresa a un mundo muy distinto al que abandonó, y el filme se nos presenta al principio como una comedia: un Walt Disney todavía parcialmente congelado es puesto al día por especialistas contratados para la ocasión que lo introducen en la psicodelia, en la música de The Grateful Dead y de Jimi Hendrix —Walt desarrolla un enorme interés por las letras de Robert Hunter, en las que planea basar un filme futuro—, en la música disco y en el hair metal, en el rap —Walt no entiende por qué se matan entre los cantantes, pero, dado que se trata de afroamericanos, y en virtud de su muy conocido racismo, el asunto le da igual—, en el trap y en la música del grupo belga Technotronic, en la actuación de todos los presidentes norteamericanos desde su congelamiento, en las trayectorias de Calista Flockhart, de George Clooney y de algo llamado Ke\$ha, en los reality shows, el funcionamiento de internet y de las redes sociales, en Star Wars, en el uso del teléfono móvil y en los desafíos de TikTok. (Walt se vuelve viral tras superar todos los récords existentes en el ice bucket challenge.) Quizás se trate de demasiada información para el protagonista: a veces Disney se confunde y piensa que Donald Trump es el compositor del éxito «Pump Up The Jam», salió de un reality show o es otro rapero asesinado. No recuerda si Leia Organa es hermana de Luke Skywalker o de Han Solo. Confunde «Osama» con «Obama» y no sabe si es que destruyó el World Trade Center, gobernó los Estados Unidos o promovió el turismo en Afganistán. Atribuye a Darth Vader la invención del iPhone. Etcétera. Pero Disney sigue siendo un visionario. Y se pasea por su mansión —dejando pequeños charcos a su paso, ya que su descongelamiento todavía continúa— a la espera de una idea que le permita volver a erigir un imperio. Pronto la encuentra. En un giro al melodrama, Disney descubre que tratar a los niños como adultos —una de las características más salientes de su obra anterior— ya no es tan eficaz como tratar a los adultos como niños. Y así, se embarca en la ampliación y en la reforma de Disneyland, que dos años después ya ocupa casi todo el sur de los Estados Unidos y cuenta con sedes en Bruselas, París, Nueva York, Berlín, la mayor parte de los campus universitarios estadounidenses y el barrio barcelonés de Gràcia. Pero Walt Disney nunca se detiene, y ya planea su próximo paso: la creación de Wokeland, un nuevo parque temático en el que las desgarradoras consecuencias de la existencia de la raza, la clase y el género —y de su combinatoria— serán convertidas en motivo de infinita discusión en habitaciones insonorizadas al efecto, suscitarán animadas manifestaciones por la avenida principal del complejo que conducirán de un centro comercial al siguiente y serán sólo aparentemente combatidas mediante su transformación en objetos de consumo, en camisetas y pins y libros ilustrados y artículos de opinión, que los visitantes podrán publicar en su periódico favorito —todos ellos son ya propiedad de las empresas Disney— por la tarifa publicitaria habitual y para de esa manera suscitar indignaciones digitales tan violentas como lacrimógenas y efímeras.
}

\noindent Translation: 

{\small
\noindent Title: Fragments of an invisible yesterday

Walt Disney is "brought back to life" by his descendants at a time when they, who have a difficult relationship and major differences regarding the company's future, agree to place it in the hands of its creator. Disney returns to a world very different from the one he left, and the film initially presents itself as a comedy: a Walt Disney, still partially frozen, is brought up to speed by specialists hired for the occasion. They introduce him to psychedelia, the music of The Grateful Dead and Jimi Hendrix —Walt develops a deep interest in the lyrics of Robert Hunter, on which he plans to base a future film—, disco music, hair metal, rap —Walt doesn't understand why rappers are killing each other, but since they are African American, and due to his well-known racism, he doesn’t really care—, trap music, and the sounds of the Belgian group Technotronic. He's also brought up to date on the presidencies of all the U.S. leaders since his freezing, the careers of Calista Flockhart, George Clooney, and something called Ke\$ha, reality shows, the workings of the internet and social networks, Star Wars, mobile phone use, and TikTok challenges. (Walt goes viral after breaking all existing records in the ice bucket challenge.) Perhaps this is too much information for the protagonist: at times, Disney gets confused and thinks that Donald Trump is the composer of the hit "Pump Up The Jam," that he came out of a reality show, or that he’s another murdered rapper. He can't remember if Leia Organa is Luke Skywalker's or Han Solo's sister. He confuses "Osama" with "Obama" and isn't sure whether he destroyed the World Trade Center, governed the United States, or promoted tourism in Afghanistan. He credits Darth Vader with inventing the iPhone. And so on. But Disney remains a visionary. He wanders through his mansion —leaving small puddles behind as his thawing continues— waiting for an idea that will allow him to rebuild an empire. Soon, he finds one. In a melodramatic twist, Disney discovers that treating children as adults —one of the hallmarks of his previous work— is no longer as effective as treating adults as children. Thus, he embarks on the expansion and renovation of Disneyland, which two years later already occupies most of the southern United States and has branches in Brussels, Paris, New York, Berlin, most U.S. university campuses, and the Gràcia neighborhood of Barcelona. But Walt Disney never stops, and he's already planning his next move: the creation of Wokeland, a new theme park where the harrowing consequences of race, class, and gender —and their intersections— will become the subject of endless discussions in soundproof rooms. They will spark lively protests down the main avenue of the complex, leading from one shopping mall to the next, and will only be superficially combated through their transformation into consumer products like T-shirts, pins, illustrated books, and opinion pieces that visitors can publish in their favorite newspaper —all of which are already owned by Disney— at the usual advertising rates, thus fueling digital outrage as violent as it is tear-jerking and fleeting.
}

\textbf{Best synopsis by GPT-4 in English (average score 1,43)}

\noindent\textit{Title: Don't Tell Mom}

{\small 
In the subdued light of early summer, "Don't Tell Mom" unfolds in a small, windswept town on the coast of Maine, a place where secrets are as vast as the ocean and just as treacherous. This film, woven with the threads of mystery, deep familial bonds, and the painful process of coming into one's own, promises a journey that is both haunting and profoundly moving. The story begins when the youngest of the Dumont siblings, sixteen-year-old Clara, discovers an ornate, antique key hidden beneath the floorboards of their late grandmother’s bedroom. This discovery coincurs with the first anniversary of their mother’s untimely and unexplained departure. Ania Dumont, once a vibrant artist and storyteller beloved by all, vanished without a trace, leaving her three children and a husband lost in the wake of her absence. Compelled by a mix of grief, curiosity, and the untamed hope of a child, Clara enlists the help of her older brothers, Max, a stoic and protective eighteen-year-old, and Eli, a free-spirited soul who dabbles in local folklore and the mystical, to unravel the mystery of the key. Their quest is underscored by a singular pact: Don't tell Dad. Their father, Thomas Dumont, has been a shell of his former self since Ania's disappearance. He seeks solace in the monotony of routine, a far cry from the adventurous sailor he once was, known for his daring escapades across the globe. The siblings fear that their quest might plunge him further into despair, choosing instead to tread the delicate balance of seeking the truth while shielding him from potential heartbreak. "Don't Tell Mom" is delicately paced, transitioning seamlessly between the present and snippets of the past, revealing the intricate tapestry of the Dumont family's life before the fracture. Through these glimpses, we learn of Ania’s peculiar fascination with local legends, her tales of spectral ships appearing off the coast, holding souls in limbo; of keys opening doors to other realms. The revelation of the key leads the siblings on an enigmatic journey through the town’s history, its people, and the rugged landscapes that shroud it. Along the way, they meet eccentric characters: an old sea captain who speaks in riddles, a librarian who guards ancient manuscripts like treasured heirlooms, and a reclusive artist who seems to know more about their mother than anyone. As they delve deeper, the siblings begin to uncover truths not only about their mother’s disappearance but also about themselves. Max confronts his fear of stepping into the unknown, embracing his role as the leader. Eli's beliefs in the mystical are tested, pushing the limits of his imagination and faith. Clara, meanwhile, discovers a resilience and determination she never knew she had, becoming the unifying force in their quest. The climax of "Don't Tell Mom" arrives as summer wanes, the ocean storms brewing a perfect backdrop to the torrent of revelations. The truth about Ania Dumont is poignant, a testament to the strength of maternal love and the lengths to which a mother would go to protect her children, even if it means leaving them behind. As the Dumont siblings stand together facing the relentless sea, they understand that their journey was never just about finding their mother but about uncovering the depth of bonds that unify them. In their hearts, they make an unspoken vow to keep the secret of their mother's sacrifice, ensuring that their newfound unity, built on the ruins of their past, remains unbroken. "Don't Tell Mom" is a movie about loss, resilience, and the indomitable spirit of family. It invites the audience into a world where the ordinary is intertwined with the extraordinary, where the quest for truth leads to self-discovery, and where love transcends the constraints of time and space.
}

\textbf{Best synopsis by GPT-4 in Spanish (average score 1,38)}

\noindent\textit{Title: Cuando estás en la cima no puedes caer más bajo}

{\small
En la cubierta de brillantes rascacielos, lejos de la mirada de aquellos pegados a la tierra, vive Andrés Solís. Un financiero de renombre que cree haber diseñado la vida perfecta, como si fuese su propio jardín zen, donde cada roca y cada río de arena están en el lugar exacto para proyectar calma y éxito. Pero esta armonía superficial se ve abruptamente interrumpida cuando Andrés es diagnosticado con una enfermedad crónica, incurable, que no solo desafía su percepción de invulnerabilidad sino que también le hace cuestionarse qué significa realmente estar en la cima. La película "Cuando estás en la cima no puedes caer más bajo" se sumerge en el caos emocional y existencial de un hombre que lo tenía todo y se enfrenta a la posibilidad de perderlo, no por una mala inversión o un error profesional, sino por los caprichos indescifrables del destino. A través de los ojos de Andrés, el público es llevado en un viaje desde las alturas envidiables de su éxito, a través de las profundidades de su vulnerabilidad, hasta una comprensión más profunda de lo que significa ser verdaderamente rico. La narrativa se despliega en tres actos marcados, primero, por la negación. Andrés se refugia en el trabajo, intentando ignorar su diagnóstico, convencido de que si mantiene su ritmo, su cuerpo no tendrá más remedio que seguir el paso. El segundo acto introduce la ira y la negociación, donde Andrés enfrenta a sus seres queridos, a sus médicos, e incluso a sí mismo, en una batalla por mantener el control que creía tener sobre su vida. Finalmente, llega la aceptación, pero no antes de un doloroso proceso de introspección y redescubrimiento, guiado por encuentros inesperados con personajes marginales de su vida, que funcionan como espejos distorsionados de sus propias fallas y fortalezas. La película es una exploración visualmente rica del contraste entre los paisajes urbanos fríamente futuristas, que son el reino de Andrés, y los espacios orgánicos y caóticos que comienza a frecuentar como parte de su viaje hacia la aceptación. Cámaras lentas, una paleta de colores meticulosamente seleccionada y una banda sonora ecléctica, que combina lo clásico con lo contemporáneo, sirven para subrayar el viaje emocional del protagonista y su evolución. Sin embargo, lo que distingue "Cuando estás en la cima no puedes caer más bajo" no es solo su producción estilizada ni su narrativa cautivadora, sino cómo desafía al espectador a reflexionar sobre sus propias vidas. Su verdadera genialidad radica en cómo logra ser una película profundamente personal y universal al mismo tiempo. Todos, de una forma u otra, enfrentamos o enfrentaremos pérdidas irreparables y, a través de Andrés, somos invitados a considerar cómo nuestras reacciones ante la adversidad definen quiénes somos. Andrés aprende que, tal vez, estar en la cima no tenga nada que ver con los éxitos tangibles, sino con cómo uno afronta el descenso inevitable. Descubre que aceptar nuestra propia vulnerabilidad puede abrir caminos hacia autenticidades previamente inimaginables, y que la verdadera caída no es perder lo que tenemos, sino negarnos a ver el valor en lo que queda.
}

\noindent Translation:

{\small
\noindent\textit{Title: When You're on Top, You Can't Fall Any Lower}

On the rooftop of gleaming skyscrapers, far from the gaze of those bound to the ground, lives Andrés Solís. A renowned financier who believes he has designed the perfect life, as if it were his own zen garden, where every rock and every sand river is in the exact place to project calm and success. But this superficial harmony is abruptly disrupted when Andrés is diagnosed with a chronic, incurable disease, which not only challenges his perception of invulnerability but also forces him to question what it really means to be at the top. The film "When You're on Top, You Can't Fall Any Lower" delves into the emotional and existential chaos of a man who had it all and now faces the possibility of losing it, not because of a bad investment or a professional mistake, but due to the indecipherable whims of fate. Through Andrés' eyes, the audience is taken on a journey from the enviable heights of his success, through the depths of his vulnerability, to a deeper understanding of what it truly means to be rich. The narrative unfolds in three distinct acts, first with denial. Andrés buries himself in work, trying to ignore his diagnosis, convinced that if he keeps up the pace, his body will have no choice but to follow. The second act introduces anger and bargaining, where Andrés confronts his loved ones, his doctors, and even himself, in a battle to maintain the control he thought he had over his life. Finally, acceptance arrives, but not before a painful process of introspection and rediscovery, guided by unexpected encounters with marginal characters in his life, who act as distorted mirrors of his own flaws and strengths. The film is a visually rich exploration of the contrast between the cold, futuristic urban landscapes, which are Andrés' domain, and the organic, chaotic spaces he begins to frequent as part of his journey toward acceptance. Slow cameras, a meticulously selected color palette, and an eclectic soundtrack that blends classical with contemporary music, all serve to underscore the protagonist's emotional journey and evolution. However, what sets "When You're on Top, You Can't Fall Any Lower" apart is not just its stylized production or captivating narrative, but how it challenges the viewer to reflect on their own lives. Its true brilliance lies in how it manages to be a deeply personal and universal film at the same time. We all, in one way or another, face or will face irreparable losses, and through Andrés, we are invited to consider how our reactions to adversity define who we are. Andrés learns that maybe being at the top has nothing to do with tangible successes, but with how one faces the inevitable descent. He discovers that accepting our own vulnerability can open paths to previously unimaginable authenticities and that the true fall is not losing what we have but refusing to see the value in what remains.
}

\end{document}